\newcommand{\CLASSINPUTtoptextmargin}{19mm}
\newcommand{\CLASSINPUTbottomtextmargin}{18mm}

\documentclass[conference]{IEEEtran}
\IEEEoverridecommandlockouts
\usepackage{cite}
\usepackage{amsmath,amssymb,amsfonts}
\usepackage{graphicx}
\usepackage{textcomp}
\usepackage{xcolor}

\usepackage{algorithm}
\usepackage[noend]{algpseudocode}
\usepackage{subcaption}

\def\BibTeX{{\rm B\kern-.05em{\sc i\kern-.025em b}\kern-.08em
    T\kern-.1667em\lower.7ex\hbox{E}\kern-.125emX}}
\begin{document}

\title{\vspace{6mm}ElectroVoxel: Electromagnetically Actuated Pivoting for Scalable Modular Self-Reconfigurable Robots\\
}

\author{
\IEEEauthorblockN{Martin Nisser}
\IEEEauthorblockA{\textit{MIT CSAIL} \\
Cambridge, MA \\
nisser@mit.edu}
\and
\IEEEauthorblockN{Leon Cheng}
\IEEEauthorblockA{\textit{MIT CSAIL} \\
Cambridge, MA \\
leonc@mit.edu}
\and
\IEEEauthorblockN{Yashaswini Makaram}
\IEEEauthorblockA{\textit{MIT CSAIL} \\
Cambridge, MA \\
ymakaram@mit.edu}\\
\and
\IEEEauthorblockN{Ryo Suzuki}
\IEEEauthorblockA{\textit{University of Calgary} \\
Calgary, AB \\
ryo.suzuki@ucalgary.ca}
\and
\IEEEauthorblockN{Stefanie Mueller}
\IEEEauthorblockA{\textit{MIT CSAIL} \\
Cambridge, MA \\
stefanie.mueller@mit.edu}
}

\IEEEaftertitletext{\vspace{-1.2cm}}

\maketitle

\begin{abstract}
 
This paper introduces a cube-based reconfigurable robot that utilizes an electromagnet-based actuation framework to reconfigure in three dimensions via pivoting. While a variety of actuation mechanisms for self-reconfigurable robots have been explored, they often suffer from cost, complexity, assembly and sizing requirements that prevent scaled production of such robots. To address this challenge, we use an actuation mechanism based on electromagnets embedded into the edges of each cube  to interchangeably create identically or oppositely polarized electromagnet pairs, resulting in repulsive or attractive forces, respectively. By leveraging attraction for hinge formation, and repulsion to drive pivoting maneuvers, we can reconfigure the robot by voxelizing it and actuating its constituent modules\textemdash termed \textit{Electrovoxels}\textemdash via \textit{electromagnetically actuated pivoting}. To demonstrate this, we develop fully untethered, three-dimensional self-reconfigurable robots and demonstrate 2D and 3D self-reconfiguration using pivot and traversal maneuvers on an air-table and in microgravity on a parabolic flight. This paper describes the hardware design of our robots, its pivoting framework, our reconfiguration planning software, and an evaluation of the dynamical and electrical characteristics of our system to inform the design of scalable self-reconfigurable robots.

\end{abstract}

\begin{IEEEkeywords}
modular robots; self-reconfigurable robots; space robotics
\end{IEEEkeywords}

\section{Introduction}

Roboticists have pursued a vision of modular self-reconfigurable robots for over 30 years~\cite{stoy2010self,gilpin2008miche,yim2000polybot}. Exhibiting unique benefits in adaptability, scalability, and robustness, modular self-reconfigurable robots (MSRR) promise application domains that include space exploration~\cite{yim2003modular,baca2014modred}, reconfigurable environments~\cite{sprowitz2014roombots,neubert2016soldercubes}, search and rescue~\cite{daudelin2018integrated}, and shape-changing user interfaces~\cite{roudaut2016cubimorph}. Roboticists have typically built MSRR via individually actuated modules connected by temporary joints~\cite{rus2001crystalline,sprowitz2014roombots,gilpin2010robot}. MSRR based on cubic modules have moreover achieved self-reconfigurability in two dimensions via sliding~\cite{an2008cube} and disassembly~\cite{gilpin2011making}, as well as in three dimensions via pivoting~\cite{romanishin20153d,romanishin2013m}. However, a grand challenge facing self-reconfigurable robots is their scalability~\cite{yim2007modular}---existing designs often require separate mechanisms for actuation and attachment~\cite{sprowitz2014roombots, romanishin20153d}, and in addition, require mechanical components such as motors, gears, and transmissions. However, these components are often bulky, complex, and expensive, hindering their miniaturization and scalability. 

\begin{figure}
  \centering
  \includegraphics[width=1\columnwidth]{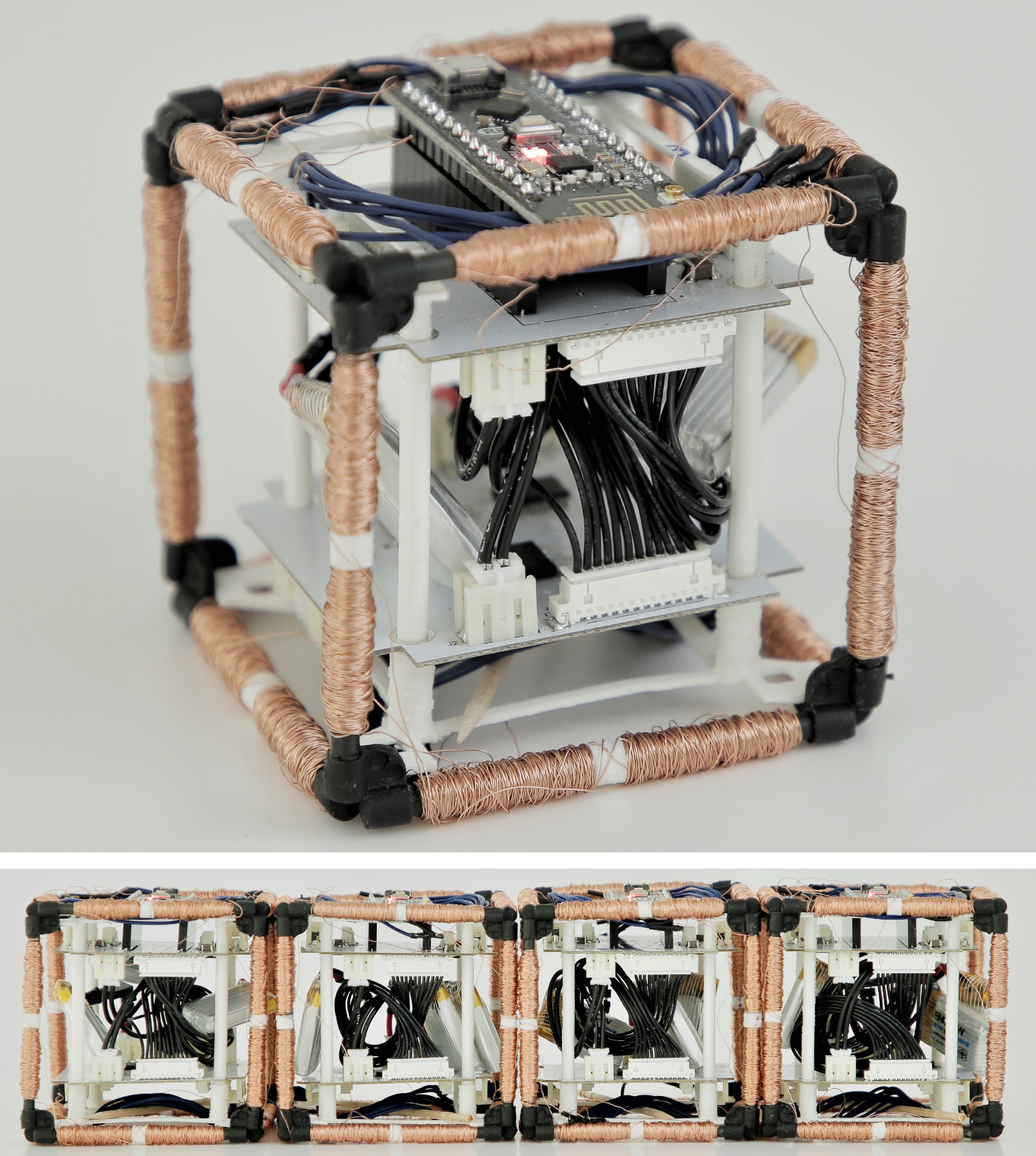}
  \\ \vspace{0.1cm}
  \resizebox{1\columnwidth}{!}{
    \includegraphics[height=1\columnwidth]{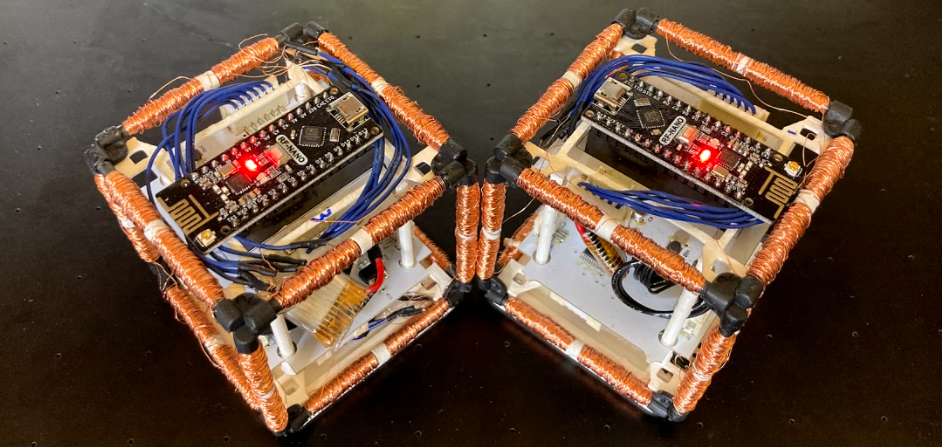}
    \hspace{0.2cm}
    \includegraphics[height=1\columnwidth]{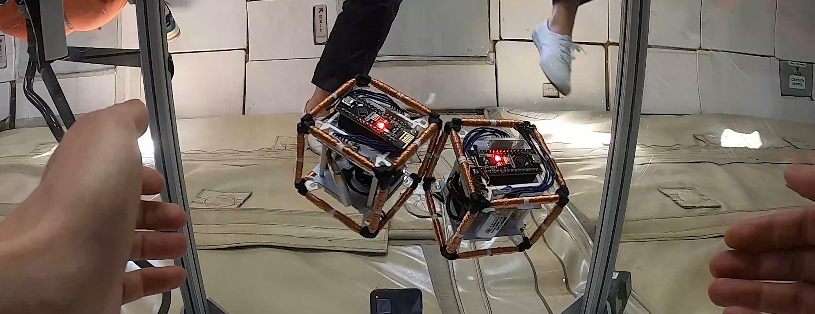}
  }
\caption{Elements of an electromagnetically actuated reconfigurable robot. (Above) A single module. (Middle) An array of four modules. (Bottom) 2D and 3D reconfiguration experiments on an air table and a microgravity environment.}
  \label{fig:TF-cube}
\vspace{-0.6cm}
\end{figure}

\begin{figure*}[t]
  \centering
  \includegraphics[width=0.99 \textwidth]{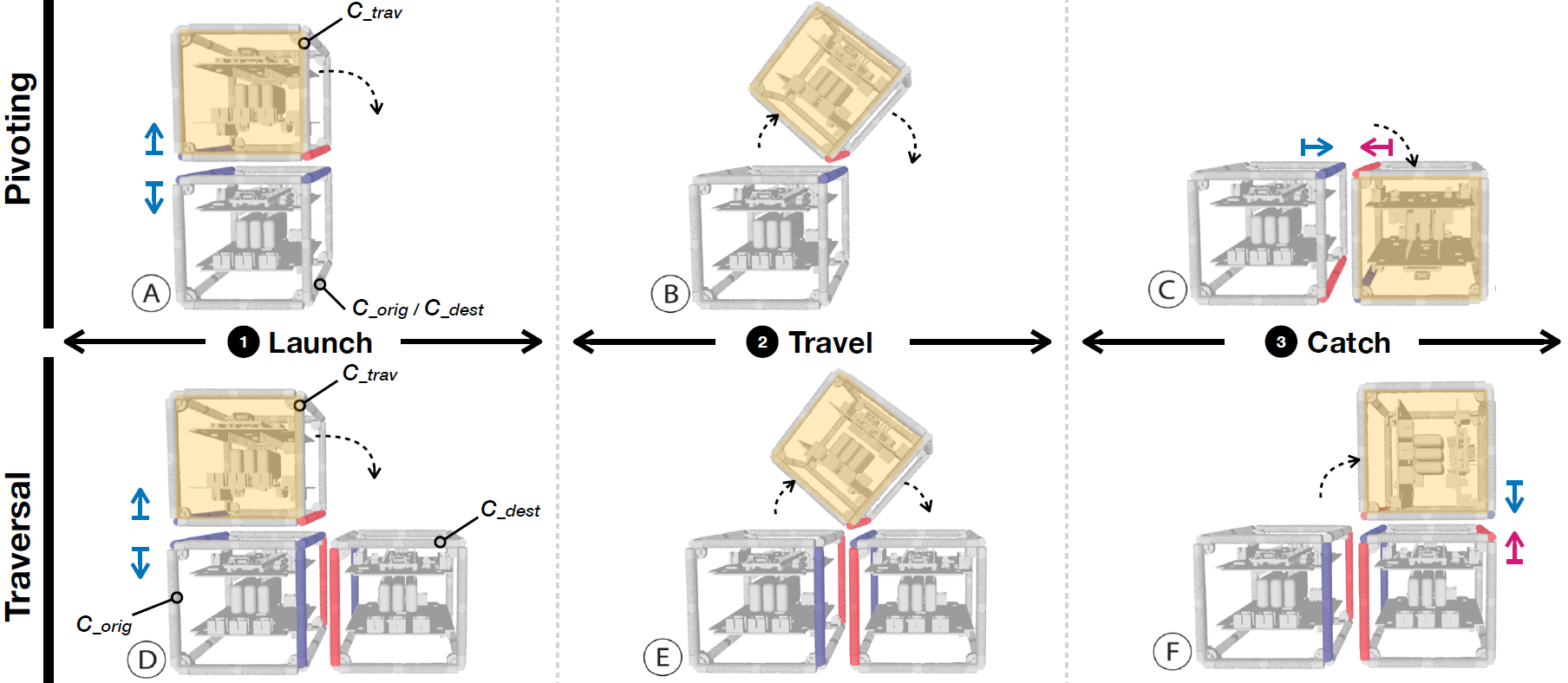}
  \caption{Reconfiguration maneuvers for (above) Pivoting and (below) Traversal. Electromagnets are shaded in red and blue to indicate polarization w.r.t. a global co-ordinate system; like polarizations repel, unlike polarizations attract.} 
  \label{fig:image-for-algo-martin}
\vspace{-0.6cm}
\end{figure*}

One promising approach to addressing these challenges is to leverage electromagnetic forces to connect and actuate modules at the same time with a single component. 
Being solid-state\textemdash that is, having no moving parts itself\textemdash it is also easy to maintain and manufacture for a large-scale system.
In particular, Nisser et al.~\cite{nisser2017electromagnetically} proposed and simulated the use of inexpensive electromagnets embedded into cube edges to actuate pivots between adjacent modules via repulsion while creating temporary hinges via attraction. Unlike traditional hinges that require mechanical attachments between two elements, this approach requires no dedicated physical mechanism and can be formed between any electromagnet pair dynamically. Kubits~\cite{hauser2020kubits} built the first physical prototype leveraging this electromagnetically actuated pivoting framework and replaced electromagnets with programmable magnets to conserve power. They also demonstrated how the electromagnetically actuated pivoting framework can generate forces significant enough to pivot modules against gravity moments if large currents are applied. However, the modules in their system were tethered to off-board electronics, and because cube edges were only partially embedded with electromagnets, interactions between all electromagnets in the system could not be observed, such as using electromagnets to maintain bonds between stationary cubes during traversals.

In this paper, we contribute the first demonstration of reconfigurable robots leveraging the electromagnetically actuated pivoting framework that  are {\it fully untethered}, supported by reconfiguration planning software and electromagnet force predictions verified experimentally. A key goal is to validate these robots' use for microgravity environments to enable near-term space industry applications~\cite{yim2003modular,yim2007modular,nisser2017electromagnetically,hauser2020kubits}, where propellant-free actuation and reconfigurability address many challenges associated with today’s limitations on launch mass and volume, as well as facilitating stowage during launch. Reconfigurable modules can enable the augmentation and replacement of structures over multiple launches, form temporary structures to aid in spacecraft inspection and astronaut assistance, function as self-sorting storage containers, and allow spacecraft to actively change their inertia properties. Microgravity alleviates demands placed on actuation forces, facilitating untethering of the modules by moving electronics onboard, and we chose electromagnet parameters such as winding number, core radius and material to limit current. We also parameterized Amp\`ere’s Force law in terms of these parameters to expand the design space of electromagnetic actuators for future modules' force and mass requirements. We simulate a microgravity environment using an air table, and deploy our modules on a parabolic flight to demonstrate untethered three-dimensional reconfigurability in space.

Relative to existing self-reconfigurable robots~\cite{romanishin2013m, romanishin20153d}, our robots are light (103g), inexpensive (\$68), and easy to fabricate (80 min/cube), promising scalability. In addition, using full assemblies, we demonstrate Sung et al.'s \cite{sung2015reconfiguration} two reconfiguration primitives, the pivot and traversal, demonstrating the electromagnetically actuated pivoting framework's compatibility with algorithms that allow reconfiguring large numbers of cube-based robots between arbitrary 3D shapes. To show how the framework complies with \cite{sung2015reconfiguration} to reconfigure more complex shapes, we constructed a web interface that simulates reconfiguration between user-defined shapes.

\section{Actuation Mechanism \& Simulation}

Reconfiguration algorithms for pivoting cubes \cite{sung2015reconfiguration,feshbach2021reconfiguring} demonstrate transitioning between arbitrarily configured 3D lattices that are provably correct in under $\mathcal{O}(n^2)$ moves, barring three inadmissible sub-configurations. These reconfiguration maneuvers rely on two reconfiguration primitives; a simple {\it pivot} between two cubes along a shared edge (Fig. \ref{fig:image-for-algo-martin}, 1st row) and a {\it traversal} between the face of one cube and another (Fig. \ref{fig:image-for-algo-martin}, 2nd row). In our work, we introduce an algorithm that describes how the electromagnetically actuated framework for individual cubes is compatible with these reconfiguration planning algorithms in three dimensions. 

\subsection{Actuation Mechanism} 
\label{sec:alg}

There are three steps to the polarization sequence for both the pivot and traversal maneuvers, which we call the \textit{Launch}, the \textit{Travel}, and \textit{Catch} phases (Fig. \ref{fig:image-for-algo-martin}). In each of these phases, three cubes are involved: a \textit{traveling} cube (the cube selected for moving), an \textit{origin} cube (from which the traveling cube launches), and a \textit{destination} cube (which catches the traveling cube). For the pivot (Fig. \ref{fig:image-for-algo-martin}, 1st row), the origin and destination cubes correspond to the same physical cubes; for the Traversal (Fig. \ref{fig:image-for-algo-martin}, 2nd row), they correspond to different cubes. The algorithm inputs are a cube ID, its desired rotation axis and rotation direction. Given these inputs, all electromagnet assignments (repulse, attract, or OFF) are uniquely defined and identified by our software.

During the Launch phase (Fig. \ref{fig:image-for-algo-martin} A and D), we polarize one electromagnet pair identically to launch the maneuver, while oppositely polarizing a second pair to form an attractive hinge. For the Traversal (Fig. \ref{fig:image-for-algo-martin}, bottom), we energize two additional pairs of electromagnets to keep the non-traveling cubes attached to one another; for this, we choose electromagnets oriented orthogonally to the launching electromagnets in order to avoid unwanted interactions between these pairs. During the Travel phase (Fig. \ref{fig:image-for-algo-martin} B and E), after a short pulse we switch the launching electromagnets off, while the remaining electromagnet pair remain attractive to maintain the hinge. During the Catch phase (Fig. \ref{fig:image-for-algo-martin} C and F), we energize a new pair of attractive electromagnets to form a stable bond in the newly acquired configuration.

\subsection{Simulation and Control Interface}

\begin{figure}[t]
  \centering
  \includegraphics[width=0.95\columnwidth]{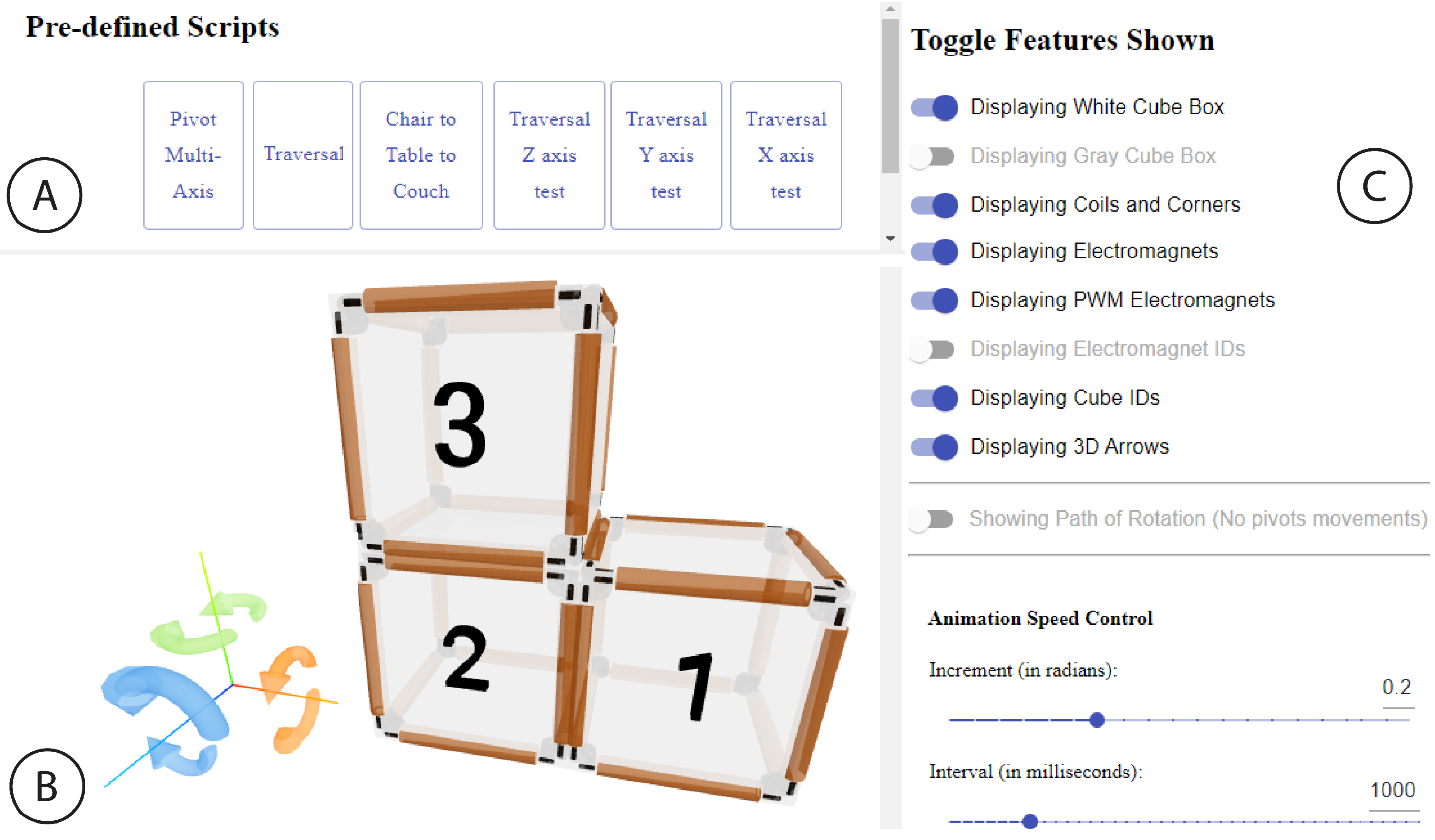}
  \caption{Web simulation for planning reconfigurations and calculating the associated electromagnet commands. (A) Pre-scripted maneuvers. (B) Viewport. (C) Settings.} 
  \label{fig:simulation-screenshot}
\vspace{-0.6cm}
\end{figure}

Manually planning pivoting maneuvers and their associated electromagnet assignments becomes intractable for more than a few cubes. To let users visualize and plan reconfiguration maneuvers, we developed a simulation (Fig. \ref{fig:simulation-screenshot}) that computes all electromagnet assignments based on desired reconfiguration maneuvers specified by the user. The simulation is browser-based and built using React, TypeScript, and Three.JS. It consists of three parts: (A) different ways to interact with the cubes (via buttons, direct manipulation, or code), (B) a viewport that simulates the cubes and affords their direct manipulation, and (C) a settings panel to toggle simulation features (e.g., different render modes for the cubes). 

We provide three ways to define maneuvers. Users can initiate maneuvers directly by clicking cubes and arrow directions, each resulting in a single pivot. Alternatively, they can launch pre-defined scripts via the buttons that encode multiple consecutive rotations. And finally, new buttons encoding different reconfiguration maneuvers can be added at any time in the underlying Typescript file. To do so, users define the number and locations of starting cubes addressed by (x,y,z) coordinates, and define each subsequent maneuver by specifying the cube number and pivot direction. 

Cubes in the viewport are affixed to a grid of cubic cells of unit length, with each cube occupying an integer address on coordinates (x,y,z). Rotations are permitted along orthogonal axes X,Y,Z, in clockwise or counter-clockwise directions. Because physical rotations require cubes to form both repulsive and attractive edges, cubes must share a face with another cube to execute a valid pivot. Each cube is assumed to have access to the local occupancy information of its neighboring cells. Prior to executing a pivot, this occupancy is checked in order to determine whether a selected rotation results in a pivot or a traversal. If the maneuver path is obstructed, the viewport returns an error message. Given a pivot direction and unobstructed path, there exists for each cube a unique valid edge about which to form a hinge, but up to two valid edges to repel and actuate the maneuver; in this case, we choose the edge corresponding to the cube with which the hinge is formed. We represent all rotations using quaternions in order to facilitate unordered rotations about multiple axes. 

Finally, the settings panel allows users to set rendering features. These include displaying IDs for cubes, their electromagnets and their polarization values, displaying occupancy requirements of neighboring cells to prevent collisions for desired maneuvers, toggling animation speed, and setting rendering fidelity. Once a sequence of maneuvers is defined, the associated electromagnet assignments can be ported to a transmitting microcontroller for deployment on the hardware.

\section{Electromagnets}

In this section, we first describe the selection of the electromagnet parameters for our system. To support further exploration of electromagnetically driven pivoting cubes, we then provide a force model to allow generating candidate electromagnets by using Amp\`ere’s Force law to compute magnetomotive force for a given electromagnet pair parametrically. We finally apply this force to a preliminary dynamical model of a 2-cube system.

\subsection{Electromagnet Parameters}

The idealized case of Amp\`ere’s law gives that an electromagnet's force is proportional to $(NI)^2\mu$, where $\mu$ is the permeability of its core, $N$ the number of turns and $I$ the current applied. To limit each cube's mass and size, we first selected the narrowest COTS ferrite cores available, at 1.625mm radius R and initial permeability $\mu_0$ of 2000. In addition, larger cores proved difficult to be supported by the rated pressure of our air table on a 60mm-side cube footprint. We next chose the COTS SMD drivers we found capable of delivering the highest current, at 1.2A continuous. Using exploratory experiments, we determined that an $N$ of 800 provided sufficient force to pivot 100g prototype cubes within interactive times of 2s. Finally, choosing a wire gauge of AWG 34 at this $NI$ yielded a coil resistance of $10.5\Omega$, allowing driving the electromagnets, microcontroller and auxiliary electronics from a single untethered power source of 11.1V-12.6V.

\subsection{Electromagnet Force Model}
The dipole approximation commonly used to find electromagnets' magnetic field strengths is invalid over short separation distances, such as when neighboring cubes are in contact. As such, we use Amp\`ere’s force law (\ref{cont}), which expresses the force $F_{1,2}$ exerted on coil 1 due to coil 2 as a double line integral over each coil's geometry where infinitesmal wire elements $\textup{d}l_1$ and $\textup{d}l_2$ in wires 1 and 2 are energized with currents $I_1$ and $I_2$, respectively, $\hat{r}_{12}$ is a unit vector from each element on wire 1 to those on wire 2 separated by distance $r$, and $\mu(I)$ is the permeability of the electromagnet core as a function of current.
  
\begin{equation}
F_{1,2} = \frac{\mu(I)}{4\pi} \int_1 \int_2 \frac{I_1 \textup{d}l_1 \times (I_2 \textup{d}l_2 \times \hat{r}_{12} )}{\left|\left|r \right|\right|^2}
\label{cont}
\end{equation}

Equation (\ref{cont}) has no known analytical solution and is discretized to give (\ref{eq_disc}), where $D_1$ and $D_2$ represent the number of discretized elements in wires 1 and 2. 

\begin{equation}
F_{1,2} = \frac{\mu(I)}{4\pi} \displaystyle\sum_{p\, = \,1}^{D_1}  \displaystyle\sum_{q\, = \,1}^{D_2} \frac{I_p \textup{d}l_p \times (I_q \textup{d}l_q \times \hat{r}_{pq} )}{\left|\left|r \right|\right|^2}
\label{eq_disc}
\end{equation}

\begin{figure}[t]
  \centering
  \includegraphics[width=0.99\columnwidth]{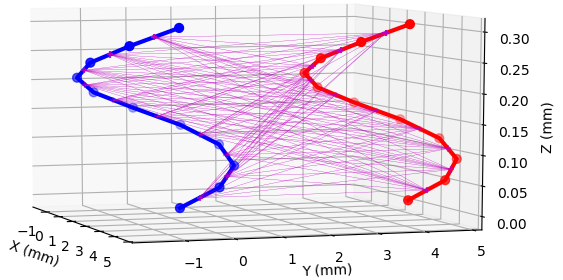}
  \caption{Computing electromagnet forces. Shown here for $D_1,D_2$=10 between 1-turn coils (force vectors in magenta).} 
  \label{fig:elec-model}
\vspace{-0.6cm}
\end{figure}

We parameterize (\ref{eq_disc}) in terms of radius, length, turns, and pitch, and solve this numerically for our electromagnet parameters, using 8000 elements ($D_1,D_2=8000$) per coil to compute $F_{1,2}$ from the sum of 64,000,000 force vectors over separation distances of 0.5mm to 20mm in 0.5mm increments (85 minutes/increment on a Razer Blade Intel(R) Core(TM) i7-8750H). Fig. \ref{fig:elec-model} illustrates this computation for single-turn electromagnet coils with 10 elements each (100 force vectors).

\subsection{Dynamic Model}

\begin{figure}[t]
  \centering
  \includegraphics[width=0.80\columnwidth]{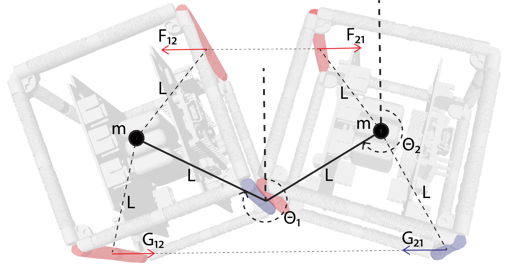}
  \caption{Dynamical model. Massless links of length L (solid lines) connect point masses m. Force F actuates the pivot via repulsion; G attracts electromagnets to form a new stable bond.} 
  \label{fig:dyna}
\vspace{-0.6cm}
\end{figure}

We apply these forces to a preliminary model (Fig. \ref{fig:dyna}). We model two 60-mm side length cubes as a 2-link pendulum consisting of two point masses $m$ placed at the distal ends of two massless links (length $L$) connecting the hinge to each cube's center of mass. Electromagnet forces are applied at locations that correspond to the positions of the other two electromagnet pairs associated with the maneuver along vectors that connect them. We derive the equations of motion and solve with Kane's Method using Python's SymPy package.

\section{Hardware}

\subsection{Electronics}

Each electromagnet is comprised of 800 turns of 34 AWG magnet wire wound around a ferromagnetic core (fair-rite 77) of 3.25mm diameter, 55.5mm length and initial permeability ($\mu_i$) of 2000, with average electrical properties characterized by a capacitance of $118.1 \mu F$, an inductance of $21.44mH$, a resistance of $10.65 \Omega$ and a Q factor of $1.265$. Each actuator (core + winding) costs just \$0.66. The circuitry for an untethered cube with 12 electromagnets consists of a microcontroller (Arduino Nano) integrated with a wireless transceiver (nRF24L01), two 16-channel GPIO expanders (Semtech SX1509) and 6 full dual H-bridges (Toshiba TB6612FNG). These are distributed evenly between two double-sided 0.78mm PCBs of square cross-section (side length 42mm) which sandwich three serially connected 3.7V batteries (ENGPOW 3.7v 150mAh Lipo, 4.2V at full charge). Combined, this allows controlling each electromagnet as to enable bidirectional pivoting in three orthogonal axes. 

We use an NRF-equipped Arduino Nano as a centralized controller to transmit commands from a laptop to modules via radio. We utilize a simple open-loop bang-bang control scheme. To accommodate addressing N cubes, each with 12 electromagnets, where each electromagnet can be polarized in two directions or turned off, each command consists of a 16-bit signed integer that encodes the cube ID $[1..N]$, electromagnet ID $[1..12]$ and its polarity $[-1,0,1]$. Individual messages are transmitted in 20 milliseconds, and separate commands can be transmitted to configure individual cubes to drive selected electromagnets using PWM at a chosen duty cycle $[0..255]$.

\subsection{Mechanical Design}
\begin{figure}[t]
\centering
\includegraphics[width=\columnwidth]{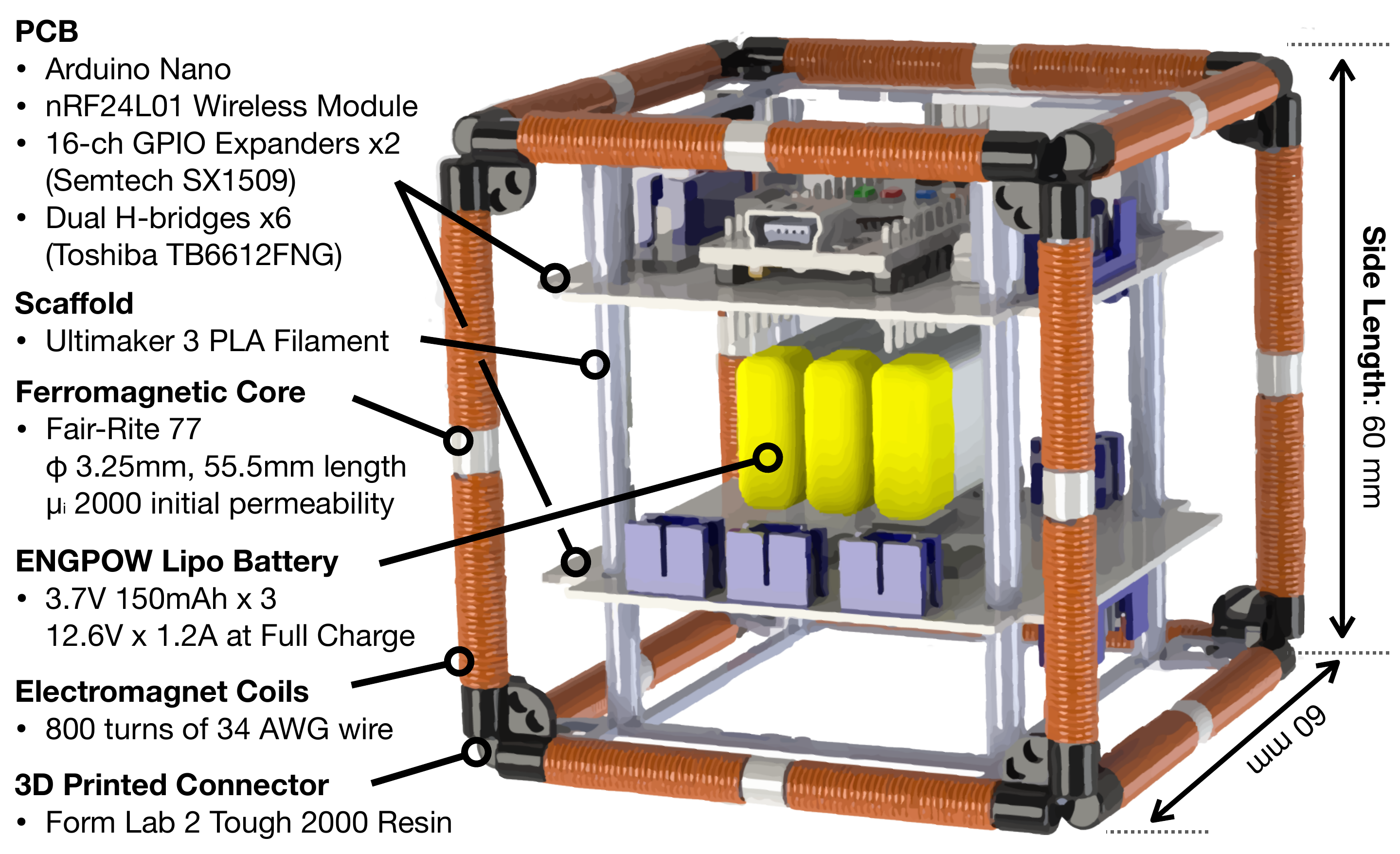}
\caption{CAD model of a cube module.}
\label{CAD}
\vspace{-0.6cm}
\end{figure}

Each module (Figures \ref{fig:TF-cube} and \ref{CAD}) is a 60mm side length cube and can be described as a unit cell of a primitive cubic Bravais lattice, with edges representing electromagnets that connect to vertices representing corner connectors. In the middle of the cube, two PCBs sandwich three batteries, centering the system's mass to limit moment of inertia, and are fixated by a scaffold that interfaces via struts to all 8 corner connectors. The corner connectors are 3D printed on a Formlabs 2 using Tough 2000 resin and the scaffold from PLA using an Ultimaker 3. Table \ref{tab1} details the cost and mass breakdown of a cube; mass and cost are for total in each row, and structure costs are based on raw material pricing.

\begin{table}[b]
\caption{Module Cost/Mass Breakdown}
\begin{center}
\begin{tabular}{|c|c|c|c|c|}
\hline
\textbf{Section} & \textbf{Part}& \textbf{Number}& \textbf{Mass (g)}& \textbf{Cost (\$)}\\
\hline
Electromagnet & Ferrite core     & 12 & 22.7 & 6.2 \\
Electromagnet & Coil winding     & 12 & 24.1 & 1.8 \\
\hline
PCB           & MCU              & 1  & 5.6  & 8.9 \\
PCB           & Cabling          & 36 & 9.2  & 11.4 \\
PCB           & Boards \& ICs    & 2  & 20.4 & 31.6 \\
PCB           & Batteries        & 3  & 15.6 & 8.2 \\
\hline
Structure     & Corners          & 8  & 2.7  & 0.4 \\
Structure     & Scaffold         & 2  & 2.8  & 0.2 \\
\hline
\textbf{Total}&     /            &  / & \textbf{103.1}& \textbf{68.7}\\
\hline
\end{tabular}
\label{tab1}
\end{center}
\end{table}

Two Molex cables connect the upper and lower PCBs, and four cables of 6 wires each (2 leads/electromagnet) harness triads of three electromagnets to the PCBs. These four triads consolidate all electromagnet wirings to two pairs of diametrically opposed corners to simplify wire routing; each triad connects three orthogonally positioned electromagnets to the PCBs in order of axis to ensure symmetry between all cubes such that they respond identically for a given command.      

Each cube took 80 minutes to assemble, discounting time to reflow-solder PCBs and manually wind electromagnets.

\section{Experiments \& Results}

\subsection{Simulation} The simulation renders pivoting maneuvers and outputs correct electromagnet IDs and polarity assignments for any valid reconfiguration; assignments were verified on hardware by reconfigurations across all dimensions and electromagnets. The simulation supports real-time interaction with up to 200 modules while rendering the associated CAD files (1.1Mb .STL), and replacing these with low resolution proxy cubes permits interaction with 1000 modules. Fig. \ref{fig:chair-to-table-to-couch} (see supplementary video) illustrates the viewport simulating and computing electromagnet assignments for reconfiguring 19 cubes from a chair to a table (via 22 maneuvers) to a couch (40 maneuvers). 

\begin{figure}[b]
  \centering
  \includegraphics[width=0.99\columnwidth]{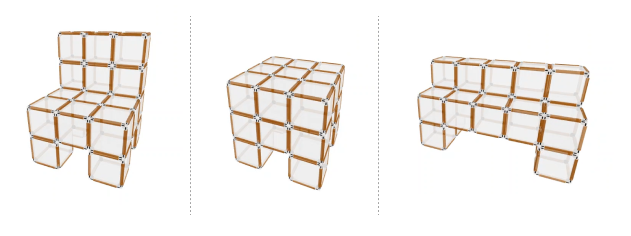}
  \caption{Reconfiguring between a chair, table and couch}
  \label{fig:chair-to-table-to-couch}
\end{figure}

\subsection{2D Experiments: Air Table}

The modules were deployed on an air table (ScienceFirst \#12000) and programmed to perform the two reconfiguration primitives; a pivot and a traversal (see supplemental video). Each electromagnet drew 11.7W (nominally 11.1V x 1.06A) to 15.1W (fully charged, 12.6V x 1.2A) for the duration of the maneuver, which was 1.53s for pivots and 1.03s for traversals.

52 pivots and 30 traversal maneuvers were performed, with a success rate of 100\% and 94\%, respectively. To yield this success rate, the electromagnets of the manually assembled cubes required careful positioning; small misalignments resulted in failures of the cubes to generate sufficient attractive forces to catch stably, and traversal maneuvers showed a higher likelihood of failure due to involving more electromagnets.

\subsection{3D Experiments: Parabolic Flight}
The modules were deployed in microgravity on a parabolic flight to observe pivoting maneuvers unimpeded by kinematic constraints from the ground plane or sliding friction. The modules were deployed in a clear 460mm side length cubic polycarbonate box with two 120mm diameter arm holes in one side for the experimenter. Ten 15-second parabolas were flown. The first 7 parabolas were utilized for calibrating the steps and timing of the experimental protocol under microgravity conditions. Most significantly, the microgravity quality was found to vary significantly between and during individual parabolas, resulting in re-programming the modules for a shortened experiment window of approximately 4 seconds of stable microgravity available before free-falling modules would impact the polycarbonate enclosure. The last 3 parabolas were used to demonstrate the pivoting maneuver between two modules.
The procedure for each 15-second parabola involved the experimenter rising from a lying to a strapped-in kneeling position on the aircraft floor, inserting hands through arm holes, and positioning the cubes while waiting for stable microgravity to settle. A trigger button was pushed to wirelessly command execution of the pivoting maneuver, and a second button was pushed to power down all cubes. 

The pivoting maneuver was executed successfully each of the three times in 1.13 seconds, while electromagnets for launch, travel and catch phases performed correctly in all tests. However, small oscillations orthogonal to the axis of rotation were observed which led to the hinge disengaging early once completing the maneuver. This result was unique to microgravity experiments and unobserved on air tables as these kinematically constrain cubes in the ground plane, and arose due to imperfect alignments including protrusions of up to 0.6mm in our manually wound coils. 

\begin{figure}
\centering
\begin{subfigure}{0.5\columnwidth}
\centering
    \includegraphics[width=0.90\columnwidth]{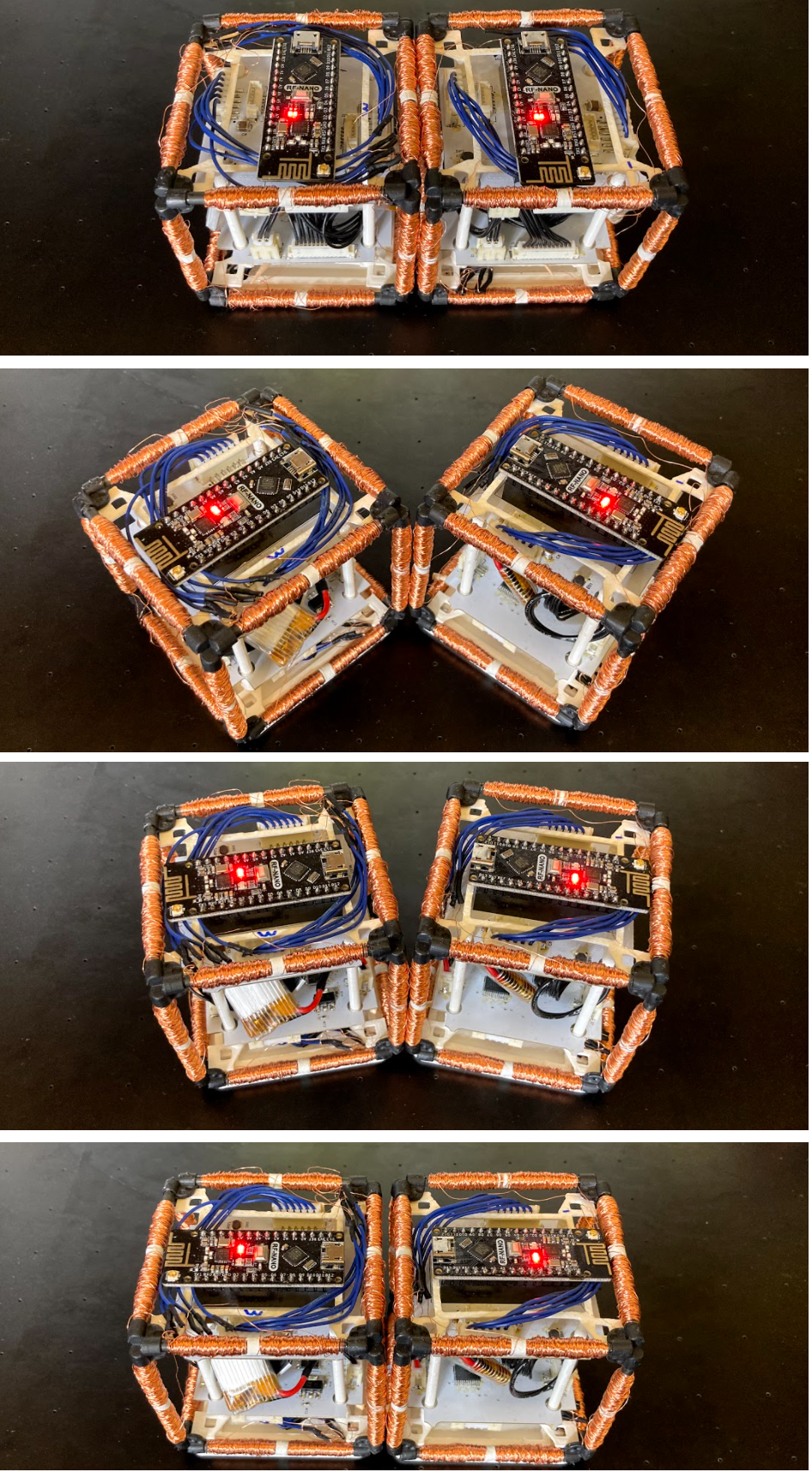}
    \caption{Air table}
    \label{fig:pivot_}
\end{subfigure}%
\begin{subfigure}{0.5\columnwidth}
\centering
    \includegraphics[width=1\columnwidth]{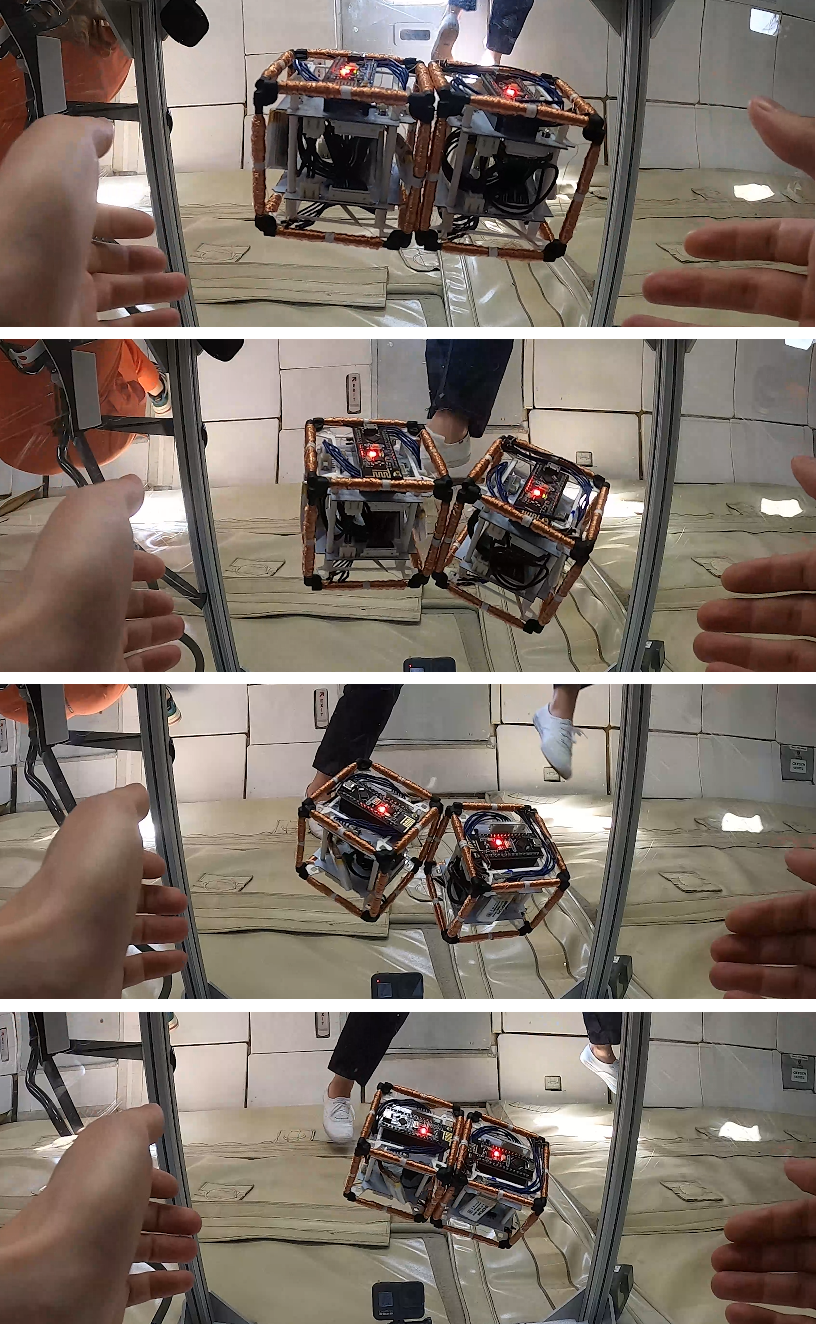}
    \caption{Parabolic flight}
    \label{fig:pivot_2}
\end{subfigure}
\caption{Two elements undergoing a pivoting maneuver (a) on an air table, a low friction surface used to simulate microgravity environments for 2D maneuvers, and (b) in microgravity aboard a parabolic flight for 3D maneuvers.  }
\label{Demos}
\vspace{-0.6cm}
\end{figure}

\subsection{Model Accuracy} 
\begin{figure}[t]
  \centering
  \includegraphics[width=0.99\columnwidth]{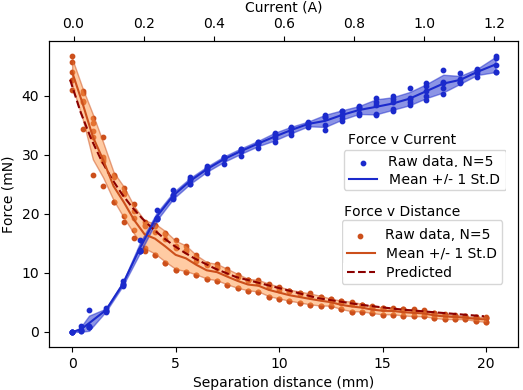}
  \caption{Force v current and electromagnet separation distance, showing raw data and mean $\pm$1 St.D for N=5. Predicted force-distance relationship from (\ref{eq_disc}) shown for comparison.}
  \label{fig:force_graph}
\vspace{-0.6cm}
\end{figure}

We measured the force generated between two electromagnets using a Mettler Toledo Precision Balance ME203T/00 (10$\mu$N precision) while varying their separation distance and their currents, shown in Fig. \ref{fig:force_graph}. We held two electromagnets at a fixed separation distance of 0.5mm, varying the current applied to each electromagnet from 0A to 1.2A in 0.05A increments. We then held the electromagnets at a fixed current of 1.2A, varying the separation distance from 0.5mm to 20mm in 0.5mm increments. Each of these experiments were conducted 5 times. Measured data are plotted in raw form (scatter plots) and as mean$\pm$1 standard deviation (shaded plots) for both experiments. Using the force-current mean at I=1.2A to extract a characterised $\mu(1.2)$ of 874, we generated predicted values for force v distance (line graph) with our force model (\ref{eq_disc}), correlating well with measured data. 

Finally, tentative results of our simple dynamical model including a simulation (supplemental video) agree qualitatively with the experiments, however as physical experiments only utilize a single force value (for $I$=$1.2A$), further work is required to capture more data to validate the model.

\section{Discussion and Future Work}

This section discusses limitations and future work for our current implementation. 
First, our cubes have to date been manually assembled and soldered, and each electromagnet hand-wound, resulting in imperfect alignments of electromagnets. While sufficient to successfully showcase the electromagnetic actuation framework on an airtable and in microgravity, Fig. \ref{fig:force_graph} reveals the degree to which these misalignments reduce attraction forces, at times disengaging the hinge. This resulted in a 6\% failure rate during air table traversals and minor instabilities in microgravity without the kinematic constraint of a ground plane; in the future, coils will be machine-wound and the assembly rigidized before deployment.

Our parameterized force model accurately predicts forces between electromagnets, supporting tailoring these actuators for different module designs. The concave downwards relationship in the force v current data (Fig. \ref{fig:force_graph}) indicates the onset of saturation of the ferrite cores in addition to potential heating effects at higher currents. Although current can be raised with higher voltage sources, this diminishes the returns on force that could potentially be achieved using greater currents; by exploiting larger cores with smaller currents, more favorable force-current relationships could be achieved with future prototypes. Nonetheless, the lack of opposing gravity moments in microgravity obviates the need for large forces, and our parabolic flight deployment required preparatory testing on an air table whose air pressure limited the total core mass.

Several avenues for future work present themselves. The success of our open-loop bang bang control suggests self-correction and robustness of the electromagnetic actuation method to small disturbances, however future work could support braking the traveling cube on impact via closed-loop control. Promising bases for control could include model-based strategies that leverage the pivoting cube model's tractable dynamics, combined with IMUs and electromagnet-based inductive sensing. To conserve power in a large scale system, future modules should embed passive permanent magnets in cube faces to form stable face-to-face bonds, or replace electromagnets with electropermanent \cite{an2008cube} or programmable \cite{hauser2020kubits} magnets. Further applications of our microgravity-adapted architecture, such as tangible or swarm user interfaces \cite{le2016zooids}, could be explored using air tables, caster wheels or low friction surfaces. Of equal interest would be to incorporate power electronics such as boost circuits for reconfiguring untethered cubes against gravity moments for terrestrial applications.

\bibliographystyle{IEEEtran}
\bibliography{references}

\end{document}